\documentclass[conference]{IEEEtran}
\ifCLASSINFOpdf
\else
\fi
\usepackage{array}

\usepackage[bookmarks=false]{hyperref}
\hypersetup{
    hidelinks,
    colorlinks=false,
    linkcolor=black, 
    citecolor=black,
    urlcolor=black
}
\usepackage{amsmath}
\usepackage{adjustbox}

\usepackage{array}
\usepackage{makecell}

\usepackage[labelfont=bf]{caption}

\usepackage{tabularx,ragged2e}

\usepackage[left=1.62cm,right=1.62cm,top=1.9cm]{geometry}

\begin{document}
%
\title{Energy Efficiency in AI for 5G and Beyond: A DeepRx Case Study}

\author{\IEEEauthorblockN{Amine Lbath}
\IEEEauthorblockA{\\
Email: amine.lbath@insa-lyon.fr}
\and
\IEEEauthorblockN{Ibtissam Labriji}
\IEEEauthorblockA{\\
Email: ibtissam.labriji@nokia-bell-labs.com}
}


%


\maketitle

\begin{abstract}
This study addresses the challenge of balancing energy efficiency with performance in AI/ML models, focusing on DeepRX, a deep learning receiver based on a fully convolutional ResNet architecture. We evaluate the energy consumption of DeepRX, considering factors including FLOPs/Watt and FLOPs/clock, and find consistency between estimated and actual energy usage, influenced by memory access patterns. The research extends to comparing energy dynamics during training and inference phases. A key contribution is the application of knowledge distillation (KD) to train a compact DeepRX \textit{student} model that emulates the performance of the \textit{teacher} model but with reduced energy consumption. We experiment with different student model sizes, optimal teacher sizes, and KD hyperparameters. Performance is measured by comparing the Bit Error Rate (BER) performance versus Signal-to-Interference \& Noise Ratio (SINR) values of the distilled model and a model trained from scratch. The distilled models demonstrate a lower error floor across SINR levels, highlighting the effectiveness of KD in achieving energy-efficient AI solutions.
\end{abstract}



%
\IEEEpeerreviewmaketitle

\section{Introduction}
In an era marked by rapid technological advancements, the telecommunications industry is leading a major transformation by increasingly using Artificial Intelligence (AI) and Machine Learning (ML). This adoption promises unprecedented levels of efficiency, enhanced performance, and an improved user experience. However, it is imperative to address the challenges that accompany this progress, particularly in terms of energy consumption and operational sustainability. Given the inherent complexity of AI/ML models, their substantial energy consumption raises concerns, especially within an industry already accountable for $2$ to $3$\% of global energy consumption ~\cite{gsmaintelligence2023going}. Additionally, the economic aspect of this problem also raises concerns, as operators allocate a substantial $15$\% to $40$\% of their operating expenses (OPEX) to energy costs~\cite{gsmaintelligence2023going}. Soaring energy prices emphasize the need to balance cost efficiency with environmental sustainability. Moreover, limitations in processing power and memory storage at the edge of the network pose significant obstacles, directly impacting the performance and potential benefits of integrating AI/ML into telecommunications infrastructure.

As we dive into these issues in more detail, it is crucial to not only understand the complexities involved but also to seek innovative solutions that can balance technological advancement with energy efficiency, cost-effectiveness, and environmental sustainability. 
In this paper, we present a variety of techniques to evaluate energy consumption of ML models, applying these methodologies to a specific case, DeepRX, a novel deep learning receiver. More specifically, we focus on the original fully convolutional ResNet-based, SIMO version of the model from ~\cite{deeprx}, to assess its energy utilization. We then derive significant insights and present these findings. Subsequently, a dedicated section explores the application of knowledge distillation to the aforementioned model. This approach aims to decrease energy consumption by training a compact \textit{student} model designed to match the performance of the original model while operating more efficiently due to its reduced size.

\section{Related Work}
The rising energy demands of advanced ML models call for a thorough assessment of their energy consumption. This necessity is driven by both environmental concerns and the pursuit of cost efficiency in various ML applications. Early steps in this direction involved defining energy efficiency in a more structured manner. Essentially, energy efficiency here refers to the amount of data that can be processed or the number of tasks that can be accomplished per unit of energy, as elaborated in \cite{Efficient_Processing_of_DNN_Sze}. Following this, several studies have focused on developing methods to estimate the energy consumption of ML models. 
The conventional approach assesses the energy usage of neural networks by examining their computational complexity. \cite{Method_estimate_energy_consumption_DNN_Sze} expanded on this by considering computational and memory costs in neural network operations, offering a broader perspective on energy use. Additionally, \cite{GARCIAMARTIN2019_Estimation_energy_consumption_ML} suggested using hardware performance counters for real-time energy consumption measurement of ML models. These efforts collectively advance the understanding and management of energy efficiency in the field of ML.

Building on these energy estimation methods, various strategies have been devised to reduce the energy consumption of Deep Learning models. During the training phase, some approaches like those presented in \cite{zeus_2022} aim to optimize the power usage of training processors. Others, such as the method in \cite{6G_AI_distibuted_2022}, enhance the efficiency of distributed learning. In the inference phase, traditional techniques like neural architecture search, pruning, and quantization are commonly employed. Recent advancements include knowledge distillation (KD) as discussed in \cite{hintonDistill}, co-optimization of neural networks and hardware \cite{admmnn_2018}, and the utilization of spiking neural networks tailored for neuromorphic chips \cite{SpikingNN_2014}. These varied techniques contribute significantly to minimizing the energy footprint of Deep Learning models across different stages of their life-cycle.
Despite advancements in the field, the importance of energy efficiency is often overlooked by ML practitioners during the development of new models. This may be partly due to a lack of clear methods for measuring energy consumption. While numerous studies, such as those referenced in \cite{Efficient_DL_survey_2021} and \cite{Efficient_Processing_of_DNN_Sze}, have delved into energy-efficient AI practices, practical implementation guides are notably rare. Our paper addresses this gap, guiding readers through the process of estimating energy consumption using various tools and metrics. We demonstrate this approach with the DeepRx model as a case study.
Furthermore, we implement KD DeepRx, illustrating how it can effectively reduce the size of the model while maintaining, its performance, thereby surpassing a model of the same size built from scratch.

\section{Methodology}
In the following, we explore how to measure the energy use of ML models. This covers the use of various tools and metrics.
\subsection{Energy Assessment}\label{sec:Energy Assessment}
\subsubsection{Components of Energy Assessment in ML}

In ML, striking a balance between \textit{energy efficiency} and \textit{accuracy} is essential for efficient data processing with low energy use. Energy efficiency indicates how well a system uses energy and is linked to overall energy consumption. Notably, this total energy consumption includes both the energy for computing and the energy needed for memory operations as illustrated by \eqref{eq:total_energy}: 
\begin{equation}\label{eq:total_energy}
    \text{Energy}_{\text{Total}} = \text{Energy}_{\text{Memory}} + \text{Energy}_{\text{Computation} \ } \cite{Efficient_Processing_of_DNN_Sze}
\end{equation}
The computation part mainly involves Multiply Accumulate (MAC) operations, whose number is dictated by the architecture of the model. On the other hand, the energy for memory tasks, which can often exceed the energy for computation, deals with managing data at different storage levels. This is influenced by the design of the model, and the underlying hardware architecture, including how data is reused and stored. For instance, the layer count and weight configuration of the model can influence energy usage. Thus, models with more layers and depth may consume more energy, even with fewer weights than simpler counterparts \cite{Eneregy_Aware_Pruning_Sze}. This underscores the importance to consider both computation and memory aspects together to really understand and improve the energy efficiency of a model.
In the following, we provide an overview of various approaches for estimating and measuring the energy consumption of ML models specifically. These methods can generally be classified into three categories: metrics, software-based measurement tools and direct physical measurement instruments, such as voltmeters. 

\subsubsection{Metrics for Energy Estimation} \label{sec:Metrics for Energy Estimation}

The first approach involves employing metrics and formulas that estimate the energy consumption of a model based on its characteristics such as the number of weights, operations, and hardware specifications including the processor power. However, these estimations are based on simplified models and often do not fully account for factors like \textit{data reuse}, which involves multiple uses of cached data to minimize costly cache accesses. While data reuse can significantly impact energy consumption, using metrics based on operation counts can still yield reasonably accurate results, as the total memory access is generally related to the number of MAC operations~\cite{MathematicalOptimizationsDL}. Below, in table \ref{table:energy-metrics}, we list out existing metrics, where $E^{c}$ represents the computation energy:

\begin{table}[!ht]
\setlength\extrarowheight{2pt} 
\begin{tabularx}{0.49\textwidth}{|>{\hsize=.55\hsize\linewidth=\hsize}X |
>{\hsize=1.45\hsize\linewidth=\hsize}X|}
\hline
\textbf{Metric} & \textbf{Parameters}\\
\hline
(0): $E^{c} = \frac{\text{Operations}}{\text{Inference}} \times \frac{1}{\frac{\text{Operations/s}}{\text{Power}}}$ \cite{Metric0}&   Operations: measured in Floating Point Operations (FLOPs), Inference: forward pass \\
\hline
(1): $E^{c} = \kappa \times N \times f^3 \times D$ \cite{Metric1}  & Where: $\kappa$ := effective switched capacitance, $N$ := number of cores, $f$ := clock frequency, $D=\frac{W}{S \times N \times f}$, with $W$ := FLOPs/inference, $S$ := FLOPs/clock \\
\hline
(2) : $E^{c} = \sum_{c}^{\text{conv layers}}E_c + \sum_{f}^{\text{FC layers}}E_f$ \cite{Metric2}&   $E_c = M_c \left(\frac{a_c}{d_0} + b_c\right)$: 
$M_c$ := (Convolutional MAC Count) = $\frac{\text{FLOPs}}{2}$, $d_0$ := number of input features, $a_c$ and $b_c$ are hardware specific parameters. 
\newline $E_f = M_f \times a_f$: $M_f$ := (Fully Connected MAC Count) = $\frac{\text{FLOPs}}{2}$, $a_f$ is a hardware specific parameter. \\
\hline
\end{tabularx}
\caption{Metrics to measure the energy consumption of ML models.}
\label{table:energy-metrics}
\end{table}

Many other metrics exists, but in practice metric (0) is preferred in the literature for its simplicity and relative accuracy. Thus we  have opted to use this metrics in our study.

\subsubsection{Tools for Energy Estimation} \label{sec:Tools for Energy Estimation}
The second approach for energy measurement involves software tools that leverage built-in processor features to evaluate energy consumption. These tools, designed to track energy use of an entire processor rather than just specific applications, monitor activities like CPU, memory, and disk usage. Notable examples include Intel's Running Average
Power Limit (RAPL) interface for Intel processors and NVIDIA's System Management Interface (nvidia-smi) for their graphics cards. In table \ref{table:energy-tools}, we compile a list of tools that are both easy to use and well-maintained.

\begin{table}[!ht]
\setlength\extrarowheight{2pt} 
\begin{tabularx}{0.49\textwidth}{|>{\hsize=.38\hsize\linewidth=\hsize}X |
>{\hsize=1.62\hsize\linewidth=\hsize}X|}
\hline
\textbf{Tool} & \textbf{Description} \\ 
\hline
\href{https://www.intel.com/content/www/us/en/developer/articles/tool/power-gadget.html}{Intel Power Gadget}  & Intel software employing RAPL for Intel processor power and energy analysis. Includes a GUI for Windows and Mac.\\ 
\hline
\href{https://arxiv.org/pdf/2002.05651.pdf}{Experiment-Impact-Tracker} & Python package for energy tracking, using Intel's RAPL for CPU/DRAM and NVIDIA's nvidia-smi for GPU.\\ 
\hline
\href{https://codecarbon.io/}{CodeCarbon} & Python package estimating the electricity and carbon usage of GPU, CPU, RAM, adjusting for regional carbon intensity, using RAPL and nvidia-smi. \\ \hline
\end{tabularx}
\caption{Overview of Energy Measurement Tools}
\label{table:energy-tools}
\end{table}

Although often hardware-specific, these software tools are almost as accurate as physical power meters and easier to use \cite{rapl_for_power_measurements_2018}. 
The aforementioned measuring software tools are designed to assess the energy consumption of the entire processor, rather than quantifying solely the energy utilized by a specific program. To collect the consumption related to running DeepRx only, we followed these steps:

\begin{enumerate}
    \item Measure the \textbf{base system power}, i.e. total power prior to running the specific program. In most cases, the power consumption remains stable over time, thus the median value is considered as the base system power. For optimal accuracy, it is crucial that no new energy-intensive program is executed afterwards, apart from the model.
    \item Measure the \textbf{absolute inference/training power}, i.e. total system-wide power with the program running.
    \item Compute the \textbf{relative inference/training power}, i.e. power exclusively linked to running the program: $\textbf{Relative Inference/Training Power} = \textbf{Absolute Program Power} - \textbf{Base System Power}$.
    \item Finally, determine the total energy consumption of the application by applying the following equation: $\mathbf{\textbf{Relative Energy} = \textbf{Relative Power} \times \textbf{Running Time}}$.
\end{enumerate}

This technique is versatile and can be effectively applied to both inference and training phases.

\subsection{Knowledge Distillation}
Following the evaluation of energy consumption using specific tools and metrics, in this section we focus on enhancing the energy efficiency of DeepRx. Our approach centers on the application of knowledge distillation (KD), a strategy proven effective in prior studies. Papers \cite{OnEfficacyOfKD}, \cite{Compact_CNN_KD},  highlight its success in boosting the energy efficiency of \textit{ResNet}-based models with minimal impact on accuracy. The effectiveness of KD in lowering energy consumption, coupled with its ease of implementation, makes it a practical choice for enhancing energy efficiency.
KD, is a technique where a smaller, \textit{student} model learns from a larger \textit{teacher} model. To enhance its own accuracy, the student aims to replicate outputs of the teacher. This replication is implemented using the distillation loss: 
\begin{equation} \label{eq:Distillation}
    \mathcal{L} = \alpha \times \mathcal{L}_{student} + \mathcal{L}_{KD} \cite{hintonDistill}
\end{equation}

The loss of the student model, $\mathcal{L}_{student}$, is regulated by $\alpha$, a parameter for balancing it against other losses. The term $\mathcal{L}_{KD} = KL(p(z_t,T), p(z_s,T)))$ represents the \textit{Kullback-Leibler} (KL) divergence, which assesses the variation between predicted output class probabilities of the teacher $p(z_t, T)$ and those of the student $p(z_s, T)$. For each class $i$, given logit $z_i$, the class probability is determined by $p(z_i, T)=\frac{\exp(\frac{z_i}{T})}{\sum_j \exp(\frac{z_i}{T})}$ \cite{hintonDistill}, with the temperature parameter $T$ acting to moderate the concentration of the distribution, it 'softens' the probability distribution, meaning the probabilities become more evenly spread. It helps the student model to learn a more nuanced representation of the output of the teacher model, rather than just replicating the most likely outcomes.
Our objective is, to achieve improved accuracy for a given model size trained using knowledge distillation, compared to training the same model from scratch, without KD. To boost DeepRx's energy efficiency through KD, we implemented the following steps:
\begin{enumerate}
    \item \textbf{Choosing the Student Model}: Selecting a size, measured in FLOPs, that minimizes energy use without significantly impacting performance, thereby maximizing the gains from distilled knowledge.
    \item \textbf{Determining the Optimal Teacher Size}: It is important to identify an appropriate teacher model size, as it significantly affects the success of KD. A too-small teacher model may not provide sufficient knowledge for the improvement of the student. Conversely, a much larger teacher model, leads to the student not having enough capacity to replicate this complex behavior.
    \item \textbf{Adjusting KD Hyperparameters}: Once teacher and student models are selected, the KD hyperparameters $\alpha$ and $T$, from training loss \eqref{eq:Distillation}, should be fine-tuned.
    \item \textbf{Exploring Additional Student Sizes}: Lastly, we expanded our research to include experiments with various student model sizes, seeking broader insights.
\end{enumerate}
The performance of our model was evaluated by calculating its Bit Error Rate performance against a range of Signal to Interference \& Noise Ratio values, $BER=f(SINR)$.
To further validate the performance of DeepRx, we also referred to the \textit{Geometric Mean BER} metric. This metric offers a comprehensive assessment by accounting for performance across a range of SINR conditions, rather than focusing on a single scenario. It is expressed as:
\begin{equation} \label{eq:valid_ber}
\text{Geometric Mean BER} = \frac{1}{N} \sum_{i=1}^{N} \log(\text{BER}_{\text{SINR}_{i}}).
 \end{equation}
$N$ represents the total number of SINR values in the dataset, and $\text{BER}_{\text{SINR}_{i}}$ represents the BER corresponding to the $i$-th SINR value. A significantly negative \textit{Geometric Mean BER} suggests low BERs across most SINR values, indicating robust model performance. Conversely, a value near zero or positive indicates higher BERs, suggesting weaker model efficacy. Although this metric does not provide detailed performance insights, it is valuable for comparing different models.

\section{Results}

In this section we consider a DeepRx model of size $30$ TFLOPs (trillion floating-point operations) as our use-case. The data is simulated using \href{https://nvlabs.github.io/sionna/}{Sionna} in a 5G MIMO scenario, using 256-QAM. The model is trained using four Nvidia A100 GPUs, with $nb. \ training \ steps = 250,000$. For the inference phase, six different processors are tested: the Intel \href{https://ark.intel.com/content/www/fr/fr/ark/products/215285/intel-xeon-gold-5320-processor-39m-cache-2-20-ghz.html}{Xeon Gold 5320}, \href{https://www.intel.com/content/www/us/en/products/sku/193388/intel-xeon-gold-5220-processor-24-75m-cache-2-20-ghz/specifications.html}{Xeon Gold 5220}, \href{https://www.intel.com/content/www/us/en/products/sku/92986/intel-xeon-processor-e52620-v4-20m-cache-2-10-ghz/specifications.html}{Xeon Gold E5-2620v4}, \href{https://ark.intel.com/content/www/fr/fr/ark/products/124968/intel-core-i78650u-processor-8m-cache-up-to-4-20-ghz.html}{I7-8650U}, the Google \href{https://coral.ai/static/files/Coral-Accelerator-Module-datasheet.pdf}{Coral Edge TPU} and finally the Nvidia \href{https://www.nvidia.com/content/geforce-gtx/GEFORCE_RTX_2080Ti_User_Guide.pdf}{RTX 20280 TI}. Besides, we consider that once deployed, DeepRx is estimated to be executed at $400 \ inferences / s$ as an uplink receiver. Finally, regarding knowledge distillation, hyperparameter values of $T \in [1,5]$ and $\alpha \in [0.1, 1e+6]$ are tested. 

\subsection{Energy estimation}\label{sec:Energy estimation}
Understanding the energy footprint of a ML model constitutes the first step in limiting it. Thus, we initially estimate the energy consumption related to a forward pass. For this purpose, we compare the energy usage of DeepRx using the aforementioned CodeCarbon tool and metric (0) as well as the aforementioned processors.

\begin{figure}[h]
    \centering
    \includegraphics[width=.8\linewidth]{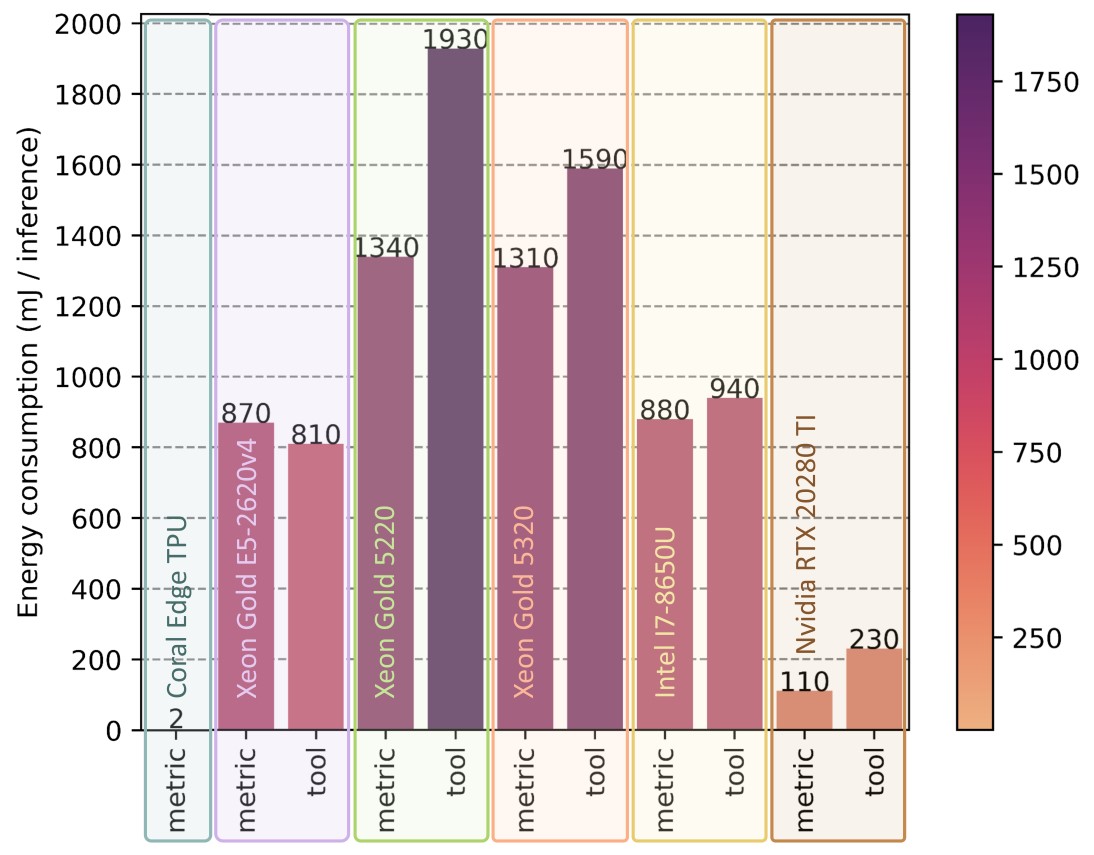}
    \caption{Inference energy usage using metric (0) vs. using tool (CodeCarbon), on different processors.}
    \label{fig:metrivs_vs_tools_infer_energy_consu_barplot}
\end{figure}
Figure \ref{fig:metrivs_vs_tools_infer_energy_consu_barplot} illustrates that the energy consumption for a single forward pass can vary significantly based on the chosen hardware. Key factors such as FLOPs/Watt and FLOPs/clock greatly influence this energy usage. For example, high-performance processors including Intel Xeon consume more energy, whereas specialized AI processors such as the Coral Edge TPU are expected to be more energy-efficient. This highlights the importance of choosing the right processor to minimize energy consumption for specific tasks.
Additionally, the comparison shows that the estimated and measured energy consumption are broadly consistent in their order of magnitude. The observed variations are partially due to memory accesses, as metric (0) does not account directly for them. Therefore, regarding our case, employing metrics as a mean to assess the energy consumption offers a credible method for acquiring an approximate evaluation.

\begin{figure}[h]
    \centering
    \includegraphics[width=\linewidth]{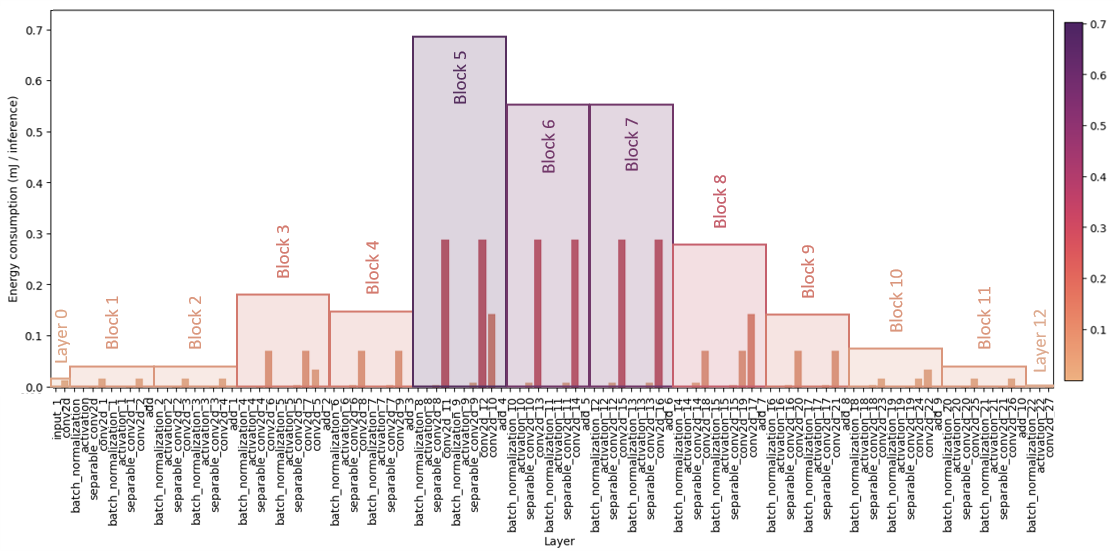}
    \caption{Block-wise and detailed layer-wise computation energy consumption per inference, on the Coral Edge TPU, using metric (0).} 
    \label{fig:layerwise_consu}
\end{figure}

The following step involves identifying the components of DeepRx that consume the most energy. Figure \eqref{fig:layerwise_consu} represents a detailed, block-wise, and layer-wise breakdown of the computation energy consumption of DeepRx per inference on the AI chip, Coral Edge TPU, using metric (0). The pattern observed in layer-wise energy consumption shows a gradual increase towards the middle layers, followed by a decrease in the final layers. Therefore, to effectively reduce the overall energy consumption of DeepRx, optimization should primarily target residual blocks $5$, $6$, and $7$, as they account for the highest energy usage. Specifically, efforts should focus on minimizing the number of parameters and computational demands of the convolutional layers within these central blocks.

To further evaluate DeepRx's energy efficiency, we compared the energy consumed during training and inference phases, depending on the number of inferences.
For training, we considered the amortized energy consumption, on a per-inference basis. This amortization distributes the training energy cost over the number of inferences the trained model will perform, thus providing a measure of energy expenditure per individual inference. The amortized training energy consumption per inference is defined by the following equation: $E_{\text{amortized, training}} = \frac{E_{\text{training}}}{n_{\text{inferences}}}$,
where $E_{\text{training}}$ is the total energy consumed during the training phase, and $n_{\text{inferences}}$ is the total number of inferences made by the model.
Subsequently, we compare this to the total energy consumption for inference, which is calculated by multiplying the energy expenditure of one inference by the number of inferences: $E_{\text{cumulated inference}} = E_{\text{single inference}} \times n_{\text{inferences}}$.
Regarding DeepRx, it is important to consider the inference energy across various inference counts to understand the efficiency of the model over its operational lifespan. For DeepRx, the inference energy on the Coral Edge TPU is $E_{\text{single inference}} = 2 \ mJ \text{ per inference}$. With one training step energy measured at $E_{\text{training step}} = 20 \ J$ and the number of training steps being $n_{\text{training steps}} = 250,000$, approximately $10^{5}$ inferences are needed to balance out the energy used in training. If DeepRx is being executed at $400 \ inferences$ per second, the energy consumed during just ten minutes of inference operations surpasses the total energy expended in the training phase. This comparison indicates that once DeepRx is deployed in a production environment, where it does not require frequent retraining, the energy used for training becomes negligible compared to the energy used for ongoing inferences. This analysis is crucial in understanding the overall energy footprint of DeepRx, emphasizing that the major energy expenditure occurs during the inference phase.

\subsection{Energy-efficiency improvement}\label{sec:Energy improvement}
Section \ref{sec:Energy estimation} highlighted that optimization efforts should be concentrated on the most energy-consuming elements, specifically the convolutional layers of the central residual blocks for the inference phase. This section provides an in-depth analysis of applying knowledge distillation to DeepRx, aiming to improve its energy efficiency. 

\begin{figure}[h]
\centering
    \includegraphics[width=.7\linewidth]{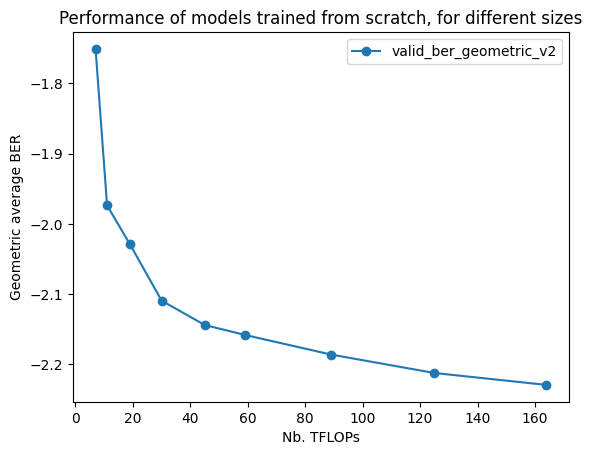}
    \caption{Performance of DeepRx models trained from scratch, for different model size (lower values indicate better performance).}
    \label{fig:DeepRx_perf_independent_different_size}
\end{figure}

The initial step involves analyzing DeepRx's performance for various model sizes when trained from scratch, prior to implementing KD. To lower the computational load of DeepRx we reduced the size of the central layers, as discussed in the previous section. Figure \ref{fig:DeepRx_perf_independent_different_size} emphasizes that, as the size of the model diminishes, the performance of DeepRx exponentially decreases. Furthermore, we can observe that an inflection point emerges at around $30$ TFLOPs, below which performance start deteriorating significantly.
Based on these findings, we focus our study on a student model of $11$ TFLOPs size. Although this scale sees some performance degradation, it is notably less than in smaller models, such as the $7$ TFLOPs variant, and KD is expected to partially mitigate this.

\begin{figure}
    \centering
    \includegraphics[width=.7\linewidth]{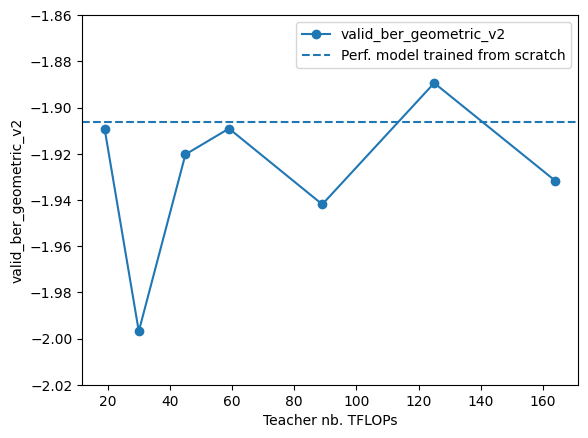}
    \caption{Performance of 11TFLOPs model trained using KD ($\alpha=500k$, $T=3$), for different teacher sizes. (Lower values indicate better performance).}
    \label{fig:DeepRx_KD_ber_different_teacher_size}
\end{figure}

After selecting the student model size, the next step involves identifying the optimal teacher model size. To achieve this, we conduct experiments covering a spectrum of teacher sizes, spanning from $19$ TFLOPs to $164$ TFLOPs. As depicted in figure \ref{fig:DeepRx_KD_ber_different_teacher_size}, DeepRx's validation performance, \textit{Geometric Mean BER}, fluctuates depending on the size of the teacher. This highlights that the performance gain associated with KD is strongly influenced by the teacher model. As discussed earlier, a teacher that is too small fails to distill useful knowledge to the student, resulting in negligible accuracy improvement. Conversely, an excessively large teacher fails to guide the student towards a better optimum, given the limited capacity of the student to replicate the intricate behavior of the teacher. This can even result in a performance drop compared to training the same model from scratch, as is the case for DeepRx trained using a teacher of size $125$ TFLOPs.
For optimal results in the DeepRx study, 
with a student model of size $11$ TFLOPs, our experiment indicates that pairing it with a $30$ TFLOPs teacher model is most effective. Based on this insight, we will maintain the use of a $30$ TFLOPs teacher model throughout the rest of our research to optimize performance gains.

\begin{table}[h]
    \centering
    \begin{tabular}{|c|c|c|c|c|}
    \hline
        $\alpha$ & $0.1$ & $50000$ & $500000$ & $1000000$ \\ \hline 
        \textit{Geometric Mean BER} & $-0.445$ & $-1.892$ & $-1.997$ & $-1.976$ \\ \hline
    \end{tabular}
\caption{Performance of $11$ TFLOPs model trained using KD ($T=3$, teacher size = $30$ TFLOPs), for different values of $\alpha$.}
\label{tab:DeepRx_KD_perf_different_alpha}    
\end{table}

After selecting the student and teacher models, it is crucial to fine-tune the $\alpha $ coefficient and the temperature $T$ hyperparameter. Consequently, we evaluated the performance of an $11$ TFLOPs DeepRx model, trained with a $30$ TFLOPs teacher, by exploring different values of $\alpha$ through grid search. 
Table \ref{tab:DeepRx_KD_perf_different_alpha} confirms that the choice of $\alpha$ significantly affects the training process. When $\alpha$ is low, the model is not able to learn exclusively from the loss $\mathcal{L}_{KD}$. This is because, soft labels of the teacher are not strict enough to properly train the student. On the other hand, if $\alpha$ is set to a very high value, the influence of $\mathcal{L}_{student}$ becomes predominant. Thus, the model loses the benefits of training using KD, effectively diminishing the potential performance gains that can be achieved. In the case of DeepRx, the optimal performance is achieved with $ \alpha=5e+5$, striking a balance between the guidance of the teacher and the learning capacity of the student.

\begin{figure}
    \centering
    \includegraphics[width=.7\linewidth]{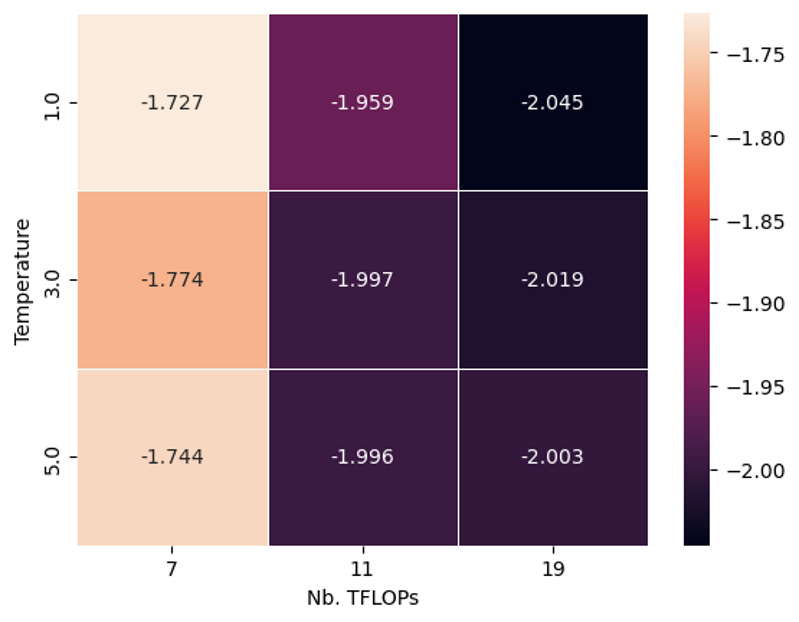}
    \caption{Performance of model trained using KD ($\alpha=500k$, teacher size = 30 TFLOPs), for different student size and different values of $T$}
    \label{fig:DeepRx_KD_perf_different_T}
\end{figure}

We further explore the impact of temperature $T$ on model performance by training student models of varying sizes through KD, using different $T$ values. 
This investigation specifically focuses on the performance of models trained using KD with an $\alpha = 500k$ and a teacher model size of $30$ TFLOPs, across various student sizes and $T$ values. Figure \ref{fig:DeepRx_KD_perf_different_T} shows the impact of the parameter $T$ on the KD distillation performance. Increasing the temperature, softens the probability distribution of the soft targets. This can be helpful up to a point, as it provides more information about the dataset than solely the hard binary targets. However, when the temperature is too high, the distribution becomes entirely uniform, losing all information about the input data. In the case of DeepRx, with an $11$ TFLOPs student and $30$ TFLOPs teacher, the optimal performance is at $T=3$.

\begin{figure}[h]
    \centering
    \includegraphics[width=0.8\linewidth]{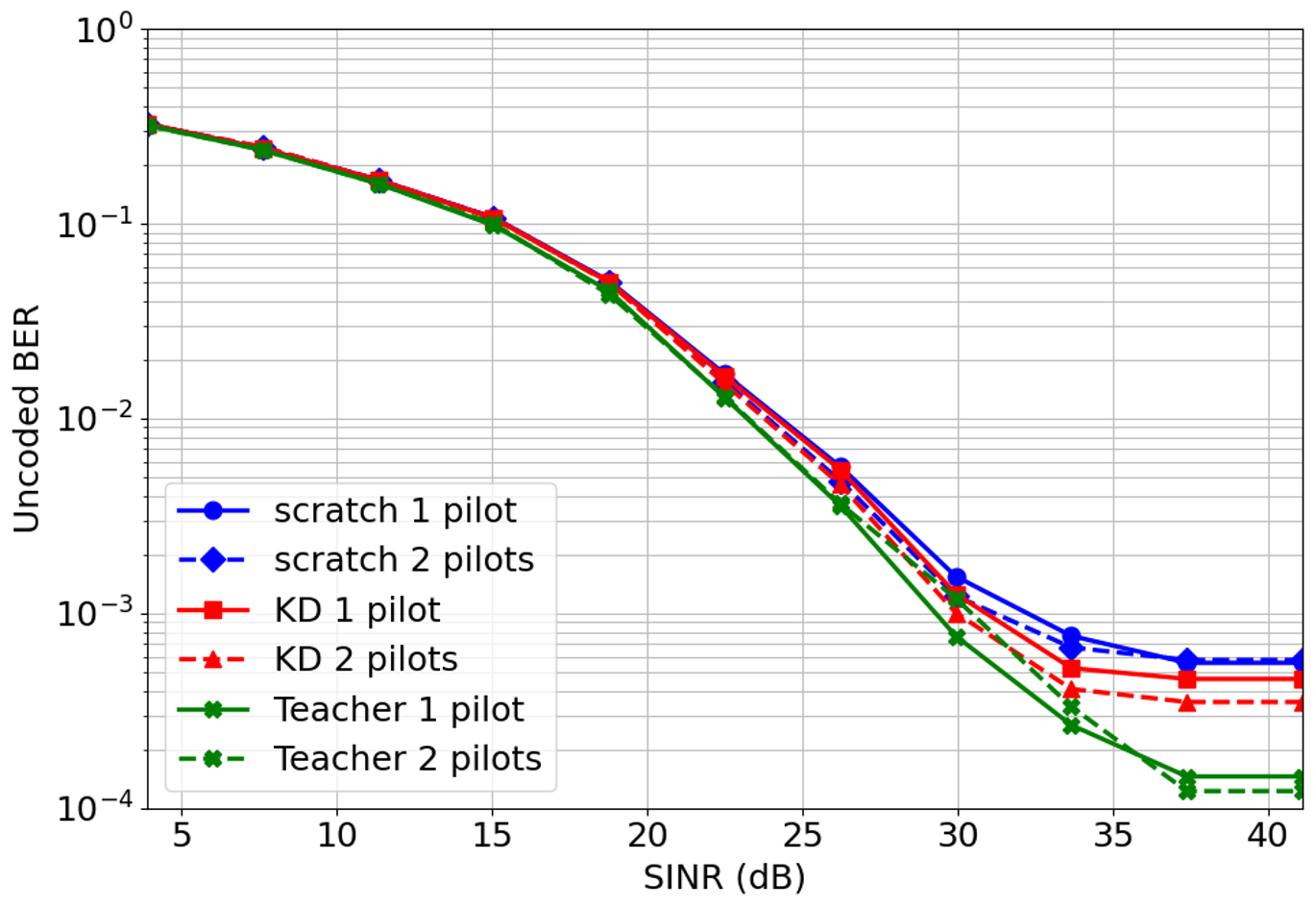}
    \caption{Performance of model trained using KD ($\alpha=500k$, teacher size = 30 TFLOPs), for different student size and different values of $T$}
    \label{fig:DeepRx_KD_BER_SINR_Perf_Compare}
\end{figure}

After a total of $700$ hours of training, the most effective approach for training an $11$ TFLOPs DeepRx model using KD involved a $30$ TFLOPs teacher model, with settings of $\alpha = 500k$ and $T=3$. To accurately assess performance enhancements, we compared the BER versus the SINR curve for this optimally trained model against that of the same model trained from scratch.
Figure \ref{fig:DeepRx_KD_BER_SINR_Perf_Compare} illustrates that models trained with KD, using an $\alpha = 500 k$ and a teacher size of $30$ TFLOPs, achieve a notably lower error floor across various SINR levels compared to counterparts trained from scratch. The graph shows that as the SINR increases, the BER for KD-trained models decreases more steeply, particularly with the model using $2$ pilots, suggesting an enhanced learning from the teacher model. This trend is consistent for different student sizes and values of $T$ indicating the robustness of KD in improving BER performance. The teacher model sets the benchmark with the lowest BER, underscoring the potential of KD in narrowing the gap towards this optimal performance, especially in high SINR scenarios.

\section{Conclusions and future directions}
We studied and implemented, through the use-case of DeepRx, techniques to improve the energy-efficiency of a ML model. We initially evaluated the detailed energy consumption of the model using metrics or tools. This helped us identify the specific component that contributes the most to the overall energy usage, throughout the life cycle of model. This insight offers direction on the specific parts of the model that require optimization efforts. It also highlighted the importance of selecting processors calibrated for the size of the model to prevent overhead. Based on previous assessments, focus was put towards enhancing the energy efficiency of DeepRx's inference phase. By applying knowledge distillation, we improved the overall performance of the model compared to the same model trained from scratch. In particular, for a $BER = 10^{-3}$, we obtained a 4 dB gain.
This promising performance improvement attained through knowledge distillation can encourage us to further pursue this approach. Further work in that direction could involve (i) training using the derivative of the teacher additionally to its logit, known as Sobolev Training, (ii) combine KD with other energy-efficient ML optimization technique, (iii) experiment an entirely different student architecture and use KD to improve gains.


\bibliography{DeepRxEnergyEfficientAI}

\begin{thebibliography}{10}
\providecommand{\url}[1]{#1}
\csname url@samestyle\endcsname
\providecommand{\newblock}{\relax}
\providecommand{\bibinfo}[2]{#2}
\providecommand{\BIBentrySTDinterwordspacing}{\spaceskip=0pt\relax}
\providecommand{\BIBentryALTinterwordstretchfactor}{4}
\providecommand{\BIBentryALTinterwordspacing}{\spaceskip=\fontdimen2\font plus
\BIBentryALTinterwordstretchfactor\fontdimen3\font minus \fontdimen4\font\relax}
\providecommand{\BIBforeignlanguage}[2]{{%
\expandafter\ifx\csname l@#1\endcsname\relax
\typeout{** WARNING: IEEEtran.bst: No hyphenation pattern has been}%
\typeout{** loaded for the language `#1'. Using the pattern for}%
\typeout{** the default language instead.}%
\else
\language=\csname l@#1\endcsname
\fi
#2}}
\providecommand{\BIBdecl}{\relax}
\BIBdecl

\bibitem{gsmaintelligence2023going}
G.~Intelligence, ``Going green: energy efficiency in telecoms,'' \emph{Product News}, 2023.

\bibitem{deeprx}
M.~Honkala, D.~Korpi, and J.~M.~J. Huttunen, ``Deeprx: Fully convolutional deep learning receiver,'' \emph{IEEE Transactions on Wireless Communications}, vol.~20, no.~6, pp. 3925--3940, 2021.

\bibitem{Efficient_Processing_of_DNN_Sze}
V.~Sze, Y.-H. Chen, T.-J. Yang, and J.~Emer, \emph{Efficient Processing of Deep Neural Networks}.\hskip 1em plus 0.5em minus 0.4em\relax Springer International Publishing, 2022.

\bibitem{Method_estimate_energy_consumption_DNN_Sze}
T.-J. Yang, Y.-H. Chen, J.~Emer, and V.~Sze, ``A method to estimate the energy consumption of deep neural networks,'' in \emph{ACSSC}.\hskip 1em plus 0.5em minus 0.4em\relax IEEE, 2017, pp. 1916--1920.

\bibitem{GARCIAMARTIN2019_Estimation_energy_consumption_ML}
E.~Garc{\'\i}a-Mart{\'\i}n, C.~F. Rodrigues, G.~Riley, and H.~Grahn, ``Estimation of energy consumption in machine learning,'' \emph{Journal of Parallel and Distributed Computing}, vol. 134, pp. 75--88, 2019.

\bibitem{zeus_2022}
J.~You, J.-W. Chung, and M.~Chowdhury, ``Zeus: Understanding and optimizing {GPU} energy consumption of {DNN} training,'' in \emph{NSDI 23}, 2023, pp. 119--139.

\bibitem{6G_AI_distibuted_2022}
M.~A. Hossain, A.~R. Hossain, and N.~Ansari, ``{AI} in {6G}: Energy-efficient distributed machine learning for multilayer heterogeneous networks,'' \emph{IEEE Network}, vol.~36, no.~6, pp. 84--91, 2022.

\bibitem{hintonDistill}
G.~Hinton, O.~Vinyals, and J.~Dean, ``Distilling the knowledge in a neural network,'' 2015.

\bibitem{admmnn_2018}
A.~Ren, T.~Zhang, S.~Ye, J.~Li, W.~Xu, X.~Qian, X.~Lin, and Y.~Wang, ``{ADMM-NN}: An algorithm-hardware co-design framework of {DNNs} using alternating direction method of multipliers,'' in \emph{ASPLOS}, 2019, pp. 925--938.

\bibitem{SpikingNN_2014}
A.~Gr{\"u}ning and S.~M. Bohte, ``Spiking neural networks: Principles and challenges,'' in \emph{ESANN}.\hskip 1em plus 0.5em minus 0.4em\relax Bruges, 2014.

\bibitem{Efficient_DL_survey_2021}
G.~Menghani, ``Efficient deep learning: A survey on making deep learning models smaller, faster, and better,'' \emph{ACM Computing Surveys}, vol.~55, no.~12, pp. 1--37, 2023.

\bibitem{Eneregy_Aware_Pruning_Sze}
V.~S. Tien-Ju~Yang, Yu-Hsin~Chen, ``Designing energy-efficient convolutional neural networks using energy-aware pruning,'' in \emph{2017 CVPR}, 2017, pp. 6071--6079.

\bibitem{MathematicalOptimizationsDL}
S.~Green, C.~M. Vineyard, and {\c{C}}.~K. Ko{\c{c}}, ``Mathematical optimizations for deep learning,'' \emph{Cyber-Physical Systems Security}, pp. 69--92, 2018.

\bibitem{Metric0}
R.~Desislavov, F.~Mart{\'\i}nez-Plumed, and J.~Hern{\'a}ndez-Orallo, ``Trends in {AI} inference energy consumption: Beyond the performance-vs-parameter laws of deep learning,'' \emph{Sustainable Computing: Informatics and Systems}, vol.~38, p. 100857, 2023.

\bibitem{Metric1}
L.~Benini and G.~d. Micheli, ``System-level power optimization: techniques and tools,'' \emph{ACM TODAES}, vol.~5, no.~2, pp. 115--192, 2000.

\bibitem{Metric2}
S.~Lahmer, A.~Khoshsirat, M.~Rossi, and A.~Zanella, ``Energy consumption of neural networks on nvidia edge boards: an empirical model,'' in \emph{WiOpt}.\hskip 1em plus 0.5em minus 0.4em\relax IEEE, 2022, pp. 365--371.

\bibitem{rapl_for_power_measurements_2018}
K.~N. Khan, M.~Hirki, T.~Niemi, J.~K. Nurminen, and Z.~Ou, ``Rapl in action: Experiences in using rapl for power measurements,'' \emph{ACM TOMPECS}, vol.~3, no.~2, pp. 1--26, 2018.

\bibitem{OnEfficacyOfKD}
J.~H. Cho and B.~Hariharan, ``On the efficacy of knowledge distillation,'' in \emph{IEEE/CVF}, 2019, pp. 4794--4802.

\bibitem{Compact_CNN_KD}
J.~Cho and M.~Lee, ``Building a compact convolutional neural network for embedded intelligent sensor systems using group sparsity and knowledge distillation,'' \emph{Sensors}, vol.~19, no.~19, 2019.

\end{thebibliography}
\bibliographystyle{IEEEtran}

\end{document}